\documentclass[11pt,a4paper]{article}
\usepackage[hyperref]{acl2017}
\usepackage{times}
\usepackage{latexsym}
\usepackage{helvet}
\usepackage{courier}
\usepackage{amsmath}
\usepackage{graphicx}
\usepackage{fancyhdr}
\usepackage{adjustbox}
\usepackage{multirow}
\usepackage{float}
\usepackage{textcomp}
\usepackage{algorithm,algorithmic}
\usepackage{amssymb,amsthm}
\usepackage{bbm}
\usepackage{url}

\aclfinalcopy 
\def\aclpaperid{3125} 

\theoremstyle{definition}

\usepackage[belowskip=-15pt,aboveskip=0pt]{caption}

\makeatletter
\newcommand{\ALOOP}[1]{\ALC@it\algorithmicloop\ #1%
    \begin{ALC@loop}}
    \newcommand{\ENDALOOP}{\end{ALC@loop}\ALC@it\algorithmicendloop}

\newcommand{\algorithmicbreak}{\textbf{break}}
\newcommand{\BREAK}{\STATE \algorithmicbreak}
\makeatother

\title{Lifelong Learning CRF for Supervised Aspect Extraction}

\author{Lei Shu, Hu Xu, Bing Liu\\
    Department of Computer Science\\
    University of Illinois at Chicago\\
    \{lshu3, hxu48, liub\}@uic.edu
}

\date{}

\begin{document}

\maketitle
\begin{abstract}
This paper makes a focused contribution to supervised aspect extraction. It shows that if the system has performed aspect extraction from many past domains and retained their results as knowledge, Conditional Random Fields (CRF) can leverage this knowledge in a lifelong learning manner to extract in a new domain markedly better than the traditional CRF without using this prior knowledge. The key innovation is that even after CRF training, the model can still improve its extraction with experiences in its applications. 

\end{abstract}

\section{Introduction}
\label{sec:introduction}
Aspect extraction is a key task of opinion mining \cite{Liu2012}. It extracts opinion targets from opinion text. For example, from the sentence ``\emph{The battery is good}'', it aims to extract ``battery'', which is a product feature, also called an {\em aspect}. 


Aspect extraction 
is commonly done using a supervised or an unsupervised approach. The unsupervised approach includes methods such as frequent pattern mining~\cite{HuL2004,PopescuNE2005,Zhu2009}, syntactic rules-based extraction \cite{ZhuangJZ2006,WangBo2008,WuZHW2009,Zhang2010,QiuLBC2011,poria2014rule}, topic modeling \cite{MeiLWSZ2007,TitovM2008,LiHuangZhu2010,Brody2010,Wang2010,Moghaddam2011,Mukherjee2012,Lin2009,ZhaoJiang2010,Jo2011,FangHuang2012ACL,WangWWW2016}, word alignment \cite{KangLiu2013IJCAI}, label propagation \cite{Zhou-wan-xiao:2013:EMNLP,shu2016lifelong}, and others~\cite{zhao2015creating}.

This paper focuses on the supervised approach \cite{Jakob2010,Choi2010,Mitchell-EtAl:2013:EMNLP} using Conditional Random Fields (CRF) \cite{Lafferty2001conditional}. It shows that the results of CRF can be significantly improved by leveraging some prior knowledge automatically mined from the extraction results of previous domains, including domains without labeled data. 
The improvement is possible because although every product (domain) is different, there is a fair amount of aspects sharing across domains~\cite{ChenLiu2014ICML}. For example, every review domain has the aspect \textit{price} and reviews of many products have the aspect \textit{battery life} or \textit{screen}. Those shared aspects may not appear in the training data but appear in unlabeled data and the test data.
We can exploit such sharing to help CRF perform much better.

Due to leveraging the knowledge gained from the past to help the new domain extraction, we are using the idea of {\em lifelong machine learning} (LML)~\cite{ChenLiu2016,thrun1998lifelong,silver2013lifelong}, which is a continuous learning paradigm that retains the knowledge learned in the past and uses it to help future learning and problem solving with possible adaptations. 

The setting of the proposed approach L-CRF ({\em Lifelong CRF}) is as follows: A CRF model $M$ has been trained with a labeled training review dataset. At a particular point in time, $M$ has extracted aspects from data in $n$ previous domains $D_1, \dots, D_n$ (which are unlabeled) and the extracted sets of aspects are $A_1, \dots, A_n$. Now, the system is faced with a new domain data $D_{n+1}$.  $M$ can leverage some {\em reliable prior knowledge} in $A_1, \dots, A_n$ to make a better extraction from $D_{n+1}$ than without leveraging this prior knowledge. 

The key innovation of L-CRF is that even after supervised training, the model can still improve its extraction in testing or its applications with experiences. Note that L-CRF is different from semi-supervised learning \cite{zhu2005semi} as the $n$ previous (unlabeled) domain data used in extraction are not used or not available during model training.

There are prior LML works for aspect extraction \cite{ChenZhiyuan2014ACL,liu2016improving}, but they were all unsupervised methods. Supervised LML methods exist \cite{chen2015lifelong,ruvolo2013ella}, but they are for classification rather than for sequence learning or labeling like CRF. A semi-supervised LML method is used in NELL\cite{mitchell2015}, but it is heuristic pattern-based. It doesn't use sequence learning and is not for aspect extraction. LML is related to transfer learning and multi-task learning~\cite{pan2010survey}, but they are also quite different (see~\cite{ChenLiu2016} for details).


To the best of our knowledge, this is the first paper that uses LML to help a supervised extraction method to markedly improve its results.  

\section{Conditional Random Fields}
CRF learns from an observation sequence $\mathbf{x}$ to estimate a label sequence $\mathbf{y}$: $p(\mathbf{y}|\mathbf{x}; \boldsymbol{\theta})$, where $\boldsymbol{\theta}$ is a set of weights. 
Let $l$ be the $l$-th position in the sequence. The core parts of CRF are a set of feature functions $\mathcal {F}=\{f_h(y_{l},y_{l-1},\mathbf{x}_l)\}_{h=1}^{H}$ and their corresponding weights $\boldsymbol{\theta}=\{\theta_h\}_{h=1}^H$. 

\noindent
{\bf Feature Functions}: We use two types of feature functions (FF). 
One is \emph{Label-Label (LL)} FF:
    \begin{equation}
    \label{ll feature function}
    \resizebox{0.85\hsize}{!}{
        $f_{ij}^{\textit{LL}}(y_l,y_{l-1})=\mathbbm{1}\{y_l=i\}\mathbbm{1}\{y_{l-1}=j\} ,\forall i,j \in \mathcal{Y},$
    }
    \end{equation}
where $\mathcal{Y}$ is the set of labels, and $\mathbbm{1}\{\cdot\} $ an indicator function.
The other is \emph{Label-Word (LW)} FF:
    \begin{equation}
    \label{lw feature function}
    \resizebox{0.85\hsize}{!}{
        $f_{iv}^{\textit{LW}}(y_l,\mathbf{x}_l) = \mathbbm{1}\{y_l=i\}\mathbbm{1}\{\mathbf{x}_l=v\} ,\forall i \in \mathcal{Y}, 
        \forall v \in \mathcal{V},$
    }
    \end{equation}
where $\mathcal{V}$ is the vocabulary. This FF returns $1$ when the $l$-th word is $v$ and the $l$-th label is $v$'s specific label $i$; otherwise $0$. $\mathbf{x}_l$ is the current word, and is represented as a multi-dimensional vector. Each dimension in the vector is a feature of $\mathbf{x}_l$.  

Following the previous work in \cite{Jakob2010}, 
we use the feature set \{W, -1W, +1W, P, -1P, +1P, G\}, where W is the word and P is its POS-tag, -1W is the previous word, -1P is its POS-tag, +1W is the next word, +1P is its POS-tag, and G is the generalized dependency feature. 

Under the Label-Word FF type, we have two sub-types of FF: {\em Label-dimension} FF and {\em Label-G} FF. Label-dimension FF is for the first 6 features, and Label-G is for the G feature. 

The \emph{Label-dimension (L$d$)} FF is defined as
\begin{equation}
\label{ld feature function}
\resizebox{0.85\hsize}{!}{
    $f_{iv^d}^{\textit{L}d}(y_l,\mathbf{x}_l) = \mathbbm{1}\{y_l=i\}\mathbbm{1}\{\mathbf{x}_l^d=v^d\}, \forall i \in \mathcal{Y}, 
    \forall v^d \in \mathcal{V}^d ,$
}
\end{equation}
where $\mathcal{V}^d$ is the set of observed values in feature $d \in \{W, -1W, +1W, P, -1P, +1P\}$ and we call $\mathcal{V}^d$ feature $d$'s feature values. Eq. (\ref{ld feature function}) is a FF that returns $1$ when $\mathbf{x}_l$'s feature $d$ equals to the feature value $v^d$ and the variable $y_l$ ($l$th label) equals to the label value $i$; otherwise 0.

We describe G  and its feature function next, which also holds the key to the proposed L-CRF. 
        
\begin{table*}[ht]
    \centering
    \scalebox{0.75}{
    \begin{tabular}{ c | c| l }
        \hline
        Index & Word & Dependency Relations  \\
        \hline
        1 & The & \{\textit{(det, battery, 2, NN , The, 1, DT)} \}\\
        \hline
        2 & battery &\{\textit{(nsubj, great, 7, JJ , battery, 2, NN), (det, battery, 2, NN , The, 1, DT), (nmod, battery, 2, NN, camera, 5, NN)} \} \\
        \hline
        3 & of &\{\textit{(case, camera, 5, NN, of, 3, IN)} \}\\
        \hline
        4 & this&\{\textit{(det, camera, 5, NN, this, 4, DT)} \}\\
        \hline
        5 & camera&\{\textit{(case, camera, 5, NN, of, 3, IN), (det, camera, 5, NN, this, 4, DT), (nmod, battery, 2, NN, camera, 5, NN)} \} \\
        \hline
        6 & is&\{\textit{(cop, great, 7, JJ , is, 6, VBZ)} \}\\
        \hline
        7 & great&\{\textit{(root, ROOT, 0, VBZ, great, 7, JJ), (nsubj, great, 7, JJ , battery, 2, NN), (cop, great, 7, JJ , is, 6, VBZ)} \} \\
        \hline 
    \end{tabular}
    }
    \caption{Dependency relations parsed from ``The battery of this camera is great''}
    \label{table:dr1}
\end{table*}
\section{General Dependency Feature (G)}
Feature G uses generalized dependency relations. What is interesting about this feature is that it enables L-CRF to use past knowledge in its sequence prediction at the test time in order to perform much better. This will become clear shortly. 
This feature 
takes a \emph{dependency pattern} as its value, which is generalized from dependency relations.

The general dependency feature (G) of the variable $\mathbf{x}_l$ takes a set of feature values $\mathcal{V}^{\textit{G}}$. Each feature value $v^{\textit{G}}$ is a dependency pattern. The \emph{Label-G (LG)} FF is defined as:
    \begin{equation}
    \label{lg feature function}
    \resizebox{0.85\hsize}{!}{
        $f_{iv^{\text{G}}}^{\textit{LG}}(y_l,\mathbf{x}_l) = \mathbbm{1}\{y_l=i\}\mathbbm{1}\{\mathbf{x}_l^{\textit{G}}=v^{\textit{G}}\} ,\forall i \in \mathcal{Y}, 
        \forall v^{\text{G}}\in \mathcal{V}^{\text{G}}.$
    }
    \end{equation}
Such a FF returns $1$ when the dependency feature of the variable $\mathbf{x}_l$ equals to a dependency pattern $v^{\textit{G}}$ and the variable $y_l$ equals to the label value $i$.




\subsection{Dependency Relation}
Dependency relations have been shown useful in many sentiment analysis applications \cite{Johansson2010,Jakob2010}. A dependency relation \footnote{We obtain dependency relations using Stanford CoreNLP: http://stanfordnlp.github.io/CoreNLP/.} is a quintuple-tuple:  
$\textit{(type, gov, govpos, dep, deppos)},$
where $\textit{type}$ is the type of the dependency relation, \textit{gov} is the \emph{governor word}, \textit{govpos} is the POS tag of the governor word, \textit{dep} is the \emph{dependent word}, and \textit{deppos} is the POS tag of the dependent word. The $l$-th word can either be the governor word or the dependent word in a dependency relation.



\subsection{Dependency Pattern}
We generalize dependency relations into {\em dependency patterns} using the following steps: 

\begin{enumerate}
    \item For each dependency relation, replace the current word (governor word or dependent word) and its POS tag with a wildcard since we already have the word (W) and the POS tag (P) features.

    \item Replace the context word (the word other than the $l$-th word) in each dependency relation with a knowledge label to form a more general dependency pattern. Let the set of aspects annotated in the training data be $K^t$. If the context word in the dependency relation appears in $K^t$, we replace it with a knowledge label `A' (aspect); otherwise `O' (other).

\end{enumerate}

For example, we work on the sentence ``The battery of this camera is great.'' The dependency relations are given in Table \ref{table:dr1}. Assume the current word is ``battery,'' and ``camera'' is annotated as an aspect. The original dependency relation between ``camera'' and ``battery'' produced by a parser is (nmod, battery, NN, camera, NN). Note that we do not use the word positions in the relations in Table \ref{table:dr1}. Since the current word's information (the word itself and its POS-tag) in the dependency relation is redundant, we replace it with a wild-card. The relation becomes (nmod, *, camera, NN). Secondly, since ``camera'' is in $K^t$, we replace ``camera'' with a general label `A'. The final dependency pattern becomes (nmod,*, A, NN).


We now explain why dependency patterns can enable a CRF model to leverage the past knowledge. The key is the knowledge label `A' above, which indicates a likely aspect. Recall that our problem setting is that when we need to extract from the new domain $D_{n+1}$ using a trained CRF model $M$, we have already extracted from many previous domains $D_1, \dots, D_n$ and retained their extracted sets of aspects $A_1, \dots, A_n$. Then, we can mine reliable aspects from $A_1, \dots, A_n$ and add them in $K^t$, which enables many knowledge labels in the dependency patterns of the new data $A_{n+1}$ due to sharing of aspects across domains. This enriches the dependency pattern features, which consequently allows more aspects to be extracted from the new domain $D_{n+1}$.

\begin{algorithm}
    \caption{Lifelong Extraction of L-CRF}\label{algo:pred}
    \begin{algorithmic}[1]
        \STATE $K_p \leftarrow \emptyset$
        \ALOOP{} 
        \STATE $F \leftarrow \text{FeatureGeneration}(D_{n+1}, K)$ \label{test:1}
        \STATE $A_{n+1} \leftarrow \text{Apply-CRF-Model}(M, F)$  \label{test:2}
        \STATE $S \leftarrow S \cup \{A_{n+1}\}$
        \STATE $K_{n+1} \leftarrow \text{Frequent-Aspects-Mining}(S,\lambda)$
        \IF {$K_p = K_{n+1}$}
        \BREAK
        \ELSE
        \STATE $K \leftarrow K^t \cup K_{n+1}$\label{test:3}
        \STATE $K_p \leftarrow K_{n+1}$\label{test:4}
        \STATE $S \leftarrow S - \{A_{n+1}\}$
        \ENDIF
        \ENDALOOP
    \end{algorithmic}
\end{algorithm}
\section{The Proposed L-CRF Algorithm}

We now present the L-CRF algorithm. As the dependency patterns for the general dependency feature do not use any actual words and they can also use the prior knowledge, they are quite powerful for cross-domain extraction (the test domain is not used in training). 

Let $K$ be a set of {\em reliable aspects} mined from the aspects extracted in past domain datasets using the CRF model $M$. Note that we assume that $M$ has already been trained using some labeled training data $D^t$. Initially, $K$ is $K^t$ (the set of all annotated aspects in the training data $D^t$). The more domains $M$ has worked on, the more aspects it extracts, and the larger the set $K$ gets. When faced with a new domain $D_{n+1}$, $K$ allows the general dependency feature to generate more dependency patterns related to aspects due to more knowledge labels `A' as we explained in the previous section. Consequently, CRF has more informed features to produce better extraction results. 

L-CRF works in two phases: {\em training phase} and {\em lifelong extraction phase}. The training phase trains a CRF model $M$ using the training data $D^t$, which is the same as normal CRF training, and will not be discussed further. In the lifelong extraction phase, $M$ is used to extract aspects from coming domains ($M$ does not change and the domain data are unlabeled). All the results from the domains are retained in past aspect store $S$. At a particular time, it is assumed $M$ has been applied to $n$ past domains, and is now faced with the $n+1$ domain. L-CRF uses $M$ and reliable aspects (denoted $K_{n+1}$) mined from $S$ and $K^t$ ($K=K^t \cup K_{n+1}$) to extract from $D_{n+1}$.  Note that aspects $K_t$ from the training data are considered always reliable as they are manually labeled, thus a subset of $K$. We cannot use all extracted aspects from past domains as reliable aspects due to many extraction errors. But those aspects that appear in multiple past domains are more likely to be correct. Thus $K$ contains those frequent aspects in $S$. The lifelong extraction phase is in Algorithm \ref{algo:pred}.

\textbf{Lifelong Extraction Phase}: 
Algorithm \ref{algo:pred} performs extraction on $D_{n+1}$ iteratively. 
\begin{enumerate}
    \item It generates features ($F$) on the data $D_{n+1}$ (line \ref{test:1}), and applies the CRF model $M$ on $F$ to produce a set of aspects $A_{n+1}$ (line \ref{test:2}). 
    \item $A_{n+1}$ is added to $S$, the past aspect store. From $S$, we mine a set of frequent aspects $K_{n+1}$. The frequency threshold is $\lambda$.
    \item If $K_{n+1}$ is the same as $K_p$ from the previous iteration, the algorithm exits as no new aspects can be found. We use an iterative process because each extraction gives new results, which may increase the size of $K$, the reliable past aspects or past knowledge. The increased $K$ may produce more dependency patterns, which can enable more extractions. 
    \item Else: some additional reliable aspects are found. $M$ may extract additional aspects in the next iteration. Lines \ref{test:3} and \ref{test:4} update the two sets for the next iteration.
\end{enumerate}


     \begin{table}[t]
         \centering
            \scalebox{0.85}{
             \begin{tabular}{c|c|c|c}
                 \hline
                 {\bf Domain}&{\bf \# Sent.}& {\bf \# Asp.}& {\bf \# non-asp. words}  \\\hline
                 Computer     &536    &1173    &7675\\\hline
                 Camera     &609    &1640    &9849\\\hline
                 Router     &509    &1239    &7264\\\hline
                 Phone         &497    &980    &7478\\\hline
                 Speaker      &510    &1299    &7546\\\hline
                 DVD Player &506    &928    &7552\\\hline
                 Mp3 Player &505    &1180     &7607\\\hline
             \end{tabular}
             }
        \caption{{Annotation details of the datasets}}\label{tab:annotation} 
     \end{table}

\begin{table*}[ht]
    \centering
    \scalebox{0.95}{
    \begin{tabular}{c|c|c c c|c c c|c c c }
            \hline    
            \multicolumn{11}{c}{\textbf{Cross-Domain}} \\\hline  
            \multirow{2}{*}{Training}&\multirow{2}{*}{Testing}&\multicolumn{3}{ |c| }{CRF}& \multicolumn{3}{ |c| }{CRF+R}& \multicolumn{3}{ |c }{L-CRF} \\

            &&$\mathcal{P}$& $\mathcal{R}$&$\mathcal{F}_1$&$\mathcal{P}$& $\mathcal{R}$&$\mathcal{F}_1$
            &$\mathcal{P}$& $\mathcal{R}$&$\mathcal{F}_1$\\\hline

            $-$Computer&Computer&86.6&51.4&64.5&23.2&90.4&37.0&82.2&62.7&71.1\\\hline
            $-$Camera&Camera&84.3&48.3&61.4&21.8&86.8&34.9&81.9&60.6&69.6\\\hline
            $-$Router&Router&86.3&48.3&61.9&24.8&92.6&39.2&82.8&60.8&70.1\\\hline
            $-$Phone&Phone&72.5&50.6&59.6&20.8&81.2&33.1&70.1&59.5&64.4\\\hline
            $-$Speaker&Speaker&87.3&60.6&71.6&22.4&91.2&35.9&84.5&71.5&77.4\\\hline
            $-$DVDplayer&DVDplayer&72.7&63.2&67.6&16.4&90.7&27.7&69.7&71.5&70.6\\\hline
            $-$Mp3player&Mp3player&87.5&49.4&63.2&20.6&91.9&33.7&84.1&60.7&70.5\\\hline\hline
            &\textbf{Average}&82.5&53.1&64.3&21.4&89.3&34.5&79.3&63.9&70.5\\\hline 
  
            \multicolumn{11}{c}{\textbf{In-Domain}}\\\hline
            $-$Computer&$-$Computer&84.0&71.4&77.2&23.2&93.9&37.3&81.6&75.8&78.6\\\hline
            $-$Camera&$-$Camera&83.7&70.3&76.4&20.8&93.7&34.1&80.7&75.4&77.9\\\hline
            $-$Router&$-$Router&85.3&71.8&78.0&22.8&93.9&36.8&82.6&76.2&79.3\\\hline
            $-$Phone&$-$Phone&85.0&71.1&77.5&25.1&93.7&39.6&82.9&74.7&78.6\\\hline
            $-$Speaker&$-$Speaker&83.8&70.3&76.5&20.1&94.3&33.2&80.1&75.8&77.9\\\hline
            $-$DVDplayer&$-$DVDplayer&85.0&72.2&78.1&20.9&94.2&34.3&81.6&76.7&79.1\\\hline
            $-$Mp3player&$-$Mp3player&83.2&72.6&77.5&20.4&94.5&33.5&79.8&77.7&78.7\\\hline\hline
            &\textbf{Average}&84.3&71.4&77.3&21.9&94.0&35.5&81.3&76.0&78.6\\\hline             
        \end{tabular}
        }
    \caption{Aspect extraction results in precision, recall and F$_1$ score: Cross-Domain and In-Domain ($-$X means all except domain X)} 
    \label{tab:result} 
\end{table*}

\section{Experiments} 
    We now evaluate the proposed L-CRF method and compare with baselines.  
    
\subsection{Evaluation Datasets}
    We use two types of data for our experiments. The first type consists of seven (7) annotated benchmark review datasets from 7 domains (types of products). Since they are annotated, they are used in training and testing. The first 4 datasets are from~\cite{HuL2004}, which actually has 5 datasets from 4 domains. Since we are mainly interested in results at the domain level, we did not use one of the domain-repeated datasets. The last 3 datasets of three domains (products) are from~\cite{liu2016improving}. These datasets are used to make up our CRF training data $D^t$ and test data $D_{n+1}$. The annotation details are given in Table \ref{tab:annotation}. 
    

    
    The second type has 50 unlabeled review datasets from 50 domains or types of products~\cite{ChenLiu2014ICML}. Each dataset has 1000 reviews. They are used as the past domain data, i.e., $D_1, \dots, D_n$ ($n=50$). Since they are not labeled, they cannot be used for training or testing. 
    
    
    
\subsection{Baseline Methods} 
    We compare L-CRF with CRF. We will not compare with unsupervised methods, which have been shown improvable by lifelong learning~\cite{ChenZhiyuan2014ACL,liu2016improving}. The frequency threshold $\lambda$ in Algorithm 1 used in our experiment to judge which extracted aspects are considered reliable is empirically set to $2$.    
    
    {\bf CRF}: We use the linear chain CRF from  \footnote{https://github.com/huangzhengsjtu/pcrf/}. Note that CRF uses all features including dependency features as the proposed L-CRF but does not employ the 50 domains unlabeled data used for lifelong learning
    
    {\bf CRF+R}: It treats the reliable aspect set $K$ as a dictionary. It adds those reliable aspects in $K$ that are not extracted by CRF but are in the test data to the final results. We want to see whether incorporating $K$ into the CRF extraction through dependency patterns in L-CRF is actually needed.     

We do not compare with domain adaptation or transfer learning because domain adaption basically uses the source domain labeled data to help learning in the target domain with few or no labeled data. Our 50 domains used in lifelong learning have no labels. So they cannot help in transfer learning. Although in transfer learning, the target domain usually has a large quantity of unlabeled data, but the 50 domains are not used as the target domains in our experiments. 
    
    
\subsection{Experiment Setting} 

To compare the systems using the same training and test data, for each dataset we use 200 sentences for training and 200 sentences for testing to avoid bias towards any dataset or domain because we will combine multiple domain datasets for CRF training. We conducted both cross-domain and in-domain tests. Our problem setting is cross-domain. In-domain is used for completeness. In both cases, we assume that extraction has been done for the 50 domains. 

    {\bf Cross-domain experiments}: We combine 6 labeled domain datasets for training (1200 sentences) and test on the 7th domain (not used in training). This gives us 7 {\em cross-domain} results. This set of tests is particularly interesting as it is desirable to have the trained model used in cross-domain situations to save manual labeling effort. 
    
    {\bf In-domain experiments}: We train and test on the same 6 domains (1200 sentences for training and 1200 sentences for testing). This also gives us 7 {\em in-domain} results. 

    {\bf Evaluating Measures}: 
    We use the popular precision $\mathcal{P}$, recall $\mathcal{R}$, and $\mathcal{F}_1$-score. 
    
    \subsection{Results and Analysis}
    
    All the experiment results are given in Table \ref{tab:result}. 
    
       {\bf Cross-domain}: Each $-$X in column 1 means that domain X is not used in training. X in column 2 means that domain X is used in testing. We can see that L-CRF is markedly better than CRF and CRF+R in $\mathcal{F}_1$. CRF+R is very poor due to poor precisions, which shows treating the reliable aspects set $K$ as a dictionary isn't a good idea. 
        
        
       {\bf In-domain}: $-$X in training and test columns means that the other 6 domains are used in both training and testing (thus in-domain). We again see that L-CRF is consistently better than CRF and CRF+R in $\mathcal{F}_1$. The amount of gain is smaller. This is expected because most aspects appeared in training probably also appear in the test data as they are reviews from the same 6 products.
        
        
    
    \section{Conclusion}
    
    This paper proposed a lifelong learning method to enable CRF to leverage the knowledge gained from extraction results of previous domains (unlabeled) to improve its extraction. Experimental results showed the effectiveness of L-CRF. The current approach does not change the CRF model itself. In our future work, we plan to modify CRF so that it can consider previous extraction results as well as the knowledge in previous CRF models. 
    
\section*{Acknowledgments} 
This work was supported in part by grants from National Science Foundation (NSF) under grant no. IIS-1407927 and IIS-1650900.

    \bibliography{acl2017}
    \bibliographystyle{acl_natbib}

\end{document}